\documentclass{article}
\usepackage{arxiv}
\usepackage[utf8]{inputenc} 
\usepackage[T1]{fontenc}    
\usepackage{hyperref}       
\usepackage{url}            
\usepackage{booktabs}       
\usepackage{amsfonts}       
\usepackage{nicefrac}       
\usepackage{microtype}      
\usepackage{lipsum}		
\usepackage{graphicx}
\usepackage{natbib}
\usepackage{doi}
\usepackage{amsmath}
\usepackage{multirow}
\setcitestyle{authoryear, open={(},close={)}}
\newcommand{\beginsupplement}{%
        \setcounter{table}{0}
        \renewcommand{\thetable}{S\arabic{table}}%
        \setcounter{figure}{0}
        \renewcommand{\thefigure}{S\arabic{figure}}%
     }

\title{Small Language Models for Tabular Data}

\author{ \href{https://orcid.org/0000-0000-0000-0000}{\includegraphics[scale=0.06]{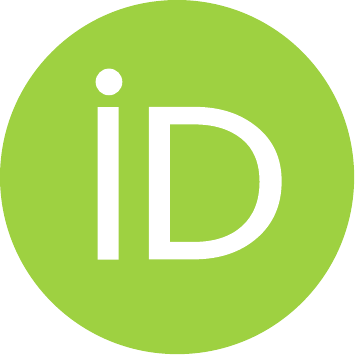}\hspace{1mm}Benjamin L. Badger}\thanks{The author would like to thank Guidehouse for support during the research and writing of this paper. Code is available at \url{https://github.com/blbadger/small-language-models}.} \\
	Guidehouse \\
	1200 19th St. NW Washington, DC 20036 \\
	\texttt{bbadger@guidehouse.com} \\
}

\date{}

\renewcommand{\headeright}{Technical Report}
\renewcommand{\undertitle}{Technical Report}

\hypersetup{
pdftitle={Small Language Models for Tabular Data},
pdfsubject={Deep Learning},
pdfauthor={Benjamin L. Badger},
pdfkeywords={Deep Learning, Representation Learning, Feature Engineering},
}

\begin{document}
\maketitle

\begin{abstract}
	Supervised deep learning is most commonly applied to difficult problems defined on large and often extensively curated datasets.  Here we demonstrate the ability of deep representation learning to address problems of classification and regression from small and poorly formed tabular datasets by encoding input information as abstracted sequences composed of a fixed number of characters per input field. We find that small models have sufficient capacity for approximation of various functions and achieve record classification benchmark accuracy.  Such models are shown to form useful embeddings of various input features in their hidden layers, even if the learned task does not explicitly require knowledge of those features. These models are also amenable to input attribution, allowing for an estimation of the importance of each input element to the model output as well as of which inputs features are effectively embedded in the model.  We present a proof-of-concept for the application of small language models to mixed tabular data without explicit feature engineering, cleaning, or preprocessing, relying on the model to perform these tasks as part of the representation learning process.
\end{abstract}


\section{Introduction}
    It has recently been observed that large deep learning models trained for natural language tasks are capable of performing certain relatively modest arithmetic operations, specifically addition and subtraction of one or two-digit integers \citep{brown2020language}.  This observation begs the question of whether a much smaller language model might be able to perform more difficult mathematical operations if trained specifically to do so.  In particular, it may be possible to perform the function approximation tasks typically undertaken using traditional machine learning techniques in which a practitioner cleans and formats a database and then assigns a set of rules with which the model chosen performs its tasks.  
    
    Such an approach would have a number of clear advantages.  A few are as follows:
    
    \begin{enumerate}
    \item Performance: the rules used for most machine learning algorithms allow for a subset of functions to be approximated, and this subset may not overlap with the set of function that one wishes to approximate.  Deep learning involves less rule assignment and allow a wider variety of functions to be approximated \citep{goodfellow2016deep}, leading to a higher likelihood that the true data generating distribution function is able to be approximated to a sufficient degree. 
    \item Less Redundancy: each addition or subtraction of an input feature to a dataset requires the practitioner to consider a number of choices as to how that input is represented, e.g. whether to use normalization for continuous variables, whether to use an embedding for high-dimensional categorical variables, whether to perform a specific type of data cleaning.  Relying instead on input representation with a fixed encoding method, addition or subtraction of input data does not change the training method or model or preprocessing steps.
    \item Flexibility:  Disparate data types may be combined with only an encoding specified.  For example, image data may be added to text and audio to make a heterogeneous dataset of characters representing a number of different sources, and again we rely on input representation in a deep learning model to determine the optimal way to represent this encoding. Moreover, the approach would be effective for a variety of output types (continuous, categorical, multi-class etc.) given one model type, whereas classical machine learning algorithms tend to perform best when applied to one output type or the other.
    \end{enumerate}
    
    The primary disadvantage of using deep learning to accomplish the cleaning and formatting tasks is that it may be relatively computationally inefficient when applied to small datasets compared to specifically-designed algorithms and traditional machine learning approaches.

    In this work we primarily investigate tasks in which the irreducible error is known (and is zero), and therefore address the ability of accurate function approximation with sequence-encoding models.  We also provide results of these models on a small benchmark dataset for reference. We focus problems that are easily expressed using a standard design matrix with rows being examples and columns being data input fields, also known as features. In the first section, data encoding methods are tested using a fixed and relatively simple model.  Next the encoding method is fixed and an alternate model choice is evaluated, and then modifications for certain datasets are explored. 
    
\subsection{Our Contribution}

    Representation of tabular data has a relatively long history for neural networks. More recently, \citep{beijbom2021} applied a pretrained DistilBERT \citep{sanh2019distilbert}, a transformer-based NLP model, to the popular Titanic dataset after converting the inputs of this dataset into sentence-like strings to create features from the input, and formed a classification on these features with an extreme gradient booster \citep{chen2016xgboost}.  Attention has also been employed on the level of a feature via TabNet \citep{arik2021tabnet}, which chooses input features to include or exclude over multiple decision steps.  Relatively simple MLPs without attention mechanisms have been found to be more effective than traditional machine learning models on various tabular benchmarks, given sufficient tuning and regularization \citep{arlind2021}. 
    
    The main point of departure for this work is that we take an optimistic view of small deep learning models' capacity and use an encoding method that may not give complete information, one that is in many ways less efficient than the standard one-hot categorical encoding or continuous direct inputs and furthermore produces a function that is much more complicated to learn.  We reason that even small deep learning models are capable of approximating an extraordinary range of functions such that increased input complexity serves to regularize the output more than anything else, by imposing a restriction of the possible functions on the encoded input that fit training data.  Results from benchmark classification tasks as well as function approximation suggest that these models are quite capable of deciphering inputs encoded as sequential characters.
    
    In light of this difference, our approach introduces a few key elements: the importance of fixed-length character encodings to avoid loss of input identity information, the notion that language models may be applied directly to contextless inputs (composed only of one-hot vectors rather than embeddings) and that these models subsequently form their own embeddings despite their modest size, the idea that input representation via deep learning can accomplish the preprocessing tasks (data cleaning, formatting, normalization etc.) normally necessary for machine learning model application, and finally that input attribution allows the experimentor to have an understanding of which inputs will be embedded in the model and which will not.
    
\section{Fixed-Length Character Encoding for Input Representation}

\subsection{Efficient Encoding versus Input Identity Loss}
    To start with, we will use a small dataset and tailor our approach to that specific dataset before then developing a generalized approach that is capable of modeling any arbitrary grid of values. Say we were given the data in tabular form as a design matrix shown in Table \ref{table:1}.  This dataset is small, has missing values, and mixed data types.
    
    \begin{table}[h!]
    \centering
    \begin{tabular}{||c c c c c c||} 
     \hline
     Market & Order Date & Time & Amount & Store Name & Total Deliverers \\ [0.5ex] 
     \hline\hline
     1 & 2015-02-06 & 22:24:17 & 1845 & Jo's & 33 \\ 
     2 & 2014-01-10 & 11:23:05 &  & Glasses R us & 16 \\
     3 &  & 21:20:40 & 1842 & Pizza & 5 \\
     \hline
    \end{tabular}
    \caption{Example Tabular Dataset}
    \label{table:1}
    \end{table}
    
    In the traditional machine learning approach to predicting some output (perhaps time until delivery), the method by which the information in this design matrix would be accessed by some model or model family depends on a number of rules, which are here referred to as preprocessing steps.  
    
    Preprocessing a dataset usually requires many decisions. For example: what should one do with missing values: assign a placeholder, or simply remove the example from our training data? What should one do if the scale of one feature is orders of magnitude different than another? Should one enforce normality on the distribution for each feature, and if so what mean and standard deviation should one apply? How should one deal with categorical inputs if only real-valued numerical vectors are accepted in the model? For categorical inputs with many possible options, should one perform dimensionality reduction to attempt to capture most of the information present in a lower-dimensional form? If so, what is the minimum number of options one should use and what method should one employ: a neural network-based embedding, sparse matrix encoding, or something else?
    
    The way to optimally represent the input data is usually theoretically unclear and computationally intractable to experimentally determine.  For example, say there were 20 features that one could either include or exclude from some classification model.  Testing all subsets of feature inclusion would require assessing $2^{20} > 1e6$ models, which is not likely to be feasible.  In this work we will bypass this concern by allowing a deep learning model to learn its own distributed representation of the input, and focus primarily on a dataset similar to that shown in Table \ref{table:1}, but with only numerical characters.  Models are also benchmarked using a dataset with letter and numerical characters. 
    
    We explore encoding the rows of an arbitrary design matrix into an input usable by a deep learning model that will then learn how to represent the input in order to best accomplish some task, usually minimization of an objective function value.  This method can be expressed as a function $f$ for our matrix as shown in Equation (\ref{eq1}).  For this work, we use $x$ to denote a specific type of encoding of an input element as a tensor, and $a_n$ to denote an input element irrespective of encoding.
    
    \begin{equation}
        x = f(1, \;2015-02-06, \; 22:24:17, \;1845, \; Pizza, \;33)
        \label{eq1}
    \end{equation}

    We approach this problem by devising a simple encoding method by which $x$ is produced without any rules specific to any particular dataset the model is applied to. This work focuses on perhaps the simplest possible encoding function $f$, a concatenation transformation on the characters of the arguments of $f$.
    
    There are a number of different functions that may be used to transform our raw input into a form suitable for a deep learning program. In the spirit of avoiding as much formatting as possible, $f$ is a one-hot character encoding followed by concatenation (and flattening if required).  All possible one-hot encodings form a basis space for each input element, and it is important that each input element is linearly independent in order to avoid inadvertently introducing prior information about those characters, as this information is almost certainly not helpful for an arbitrary task.

    Given this encoding method one notable choice remains: how to deal with missing features (column values).  This may be accomplished by simply removing training examples that exhibit missing features, and mandating that test data have some value for all features. The same approach could be applied here, but such a strategy could lead to biased results.  What happens when not only a few but the majority of examples both for training and testing are missing data on a certain feature?  Removing all examples would eliminate the majority of information in our training dataset, and is clearly not the best way to proceed.

    Perhaps the simplest way to proceed would be to assign our sequence-to-tensor mapping function $f$ to map an empty feature to an empty tensor.  For example, suppose we wanted to encode an $x_i$ which was missing the first feature value of $x_c$ (\ref{eq2}).
    
    \begin{equation}
        \begin{aligned}
        & x_c = f(1 \;	2015-02-06 \; 22:24:17 \;1845 ...) \\
        & x_i = f( \;	2015-02-06 \; 22:24:17 \;1845 ...)  
        \end{aligned}
        \label{eq2}
    \end{equation}
    
     The corresponding tensors (without concatenation and assuming 10 unique characters) are shown in Equation (\ref{eq3}) where the concatenated versions are $[0,1,0...]$ for $x_c$ and $[0,0,1,...]$ for $x_i$, resulting in a variable-length encoded set of inputs $\{x \}$.
    
    \begin{equation}
        \begin{aligned}
        & x_c = [[0,1,0,0,0,0,0,0,0,0],[0,0,1,...], ...] \\
        & x_i = [[],[0,0,1,...], ...]
        \end{aligned}
        \label{eq3}
    \end{equation}
    
\begin{figure}[h]
        \centering
        \includegraphics[width=0.9\textwidth]{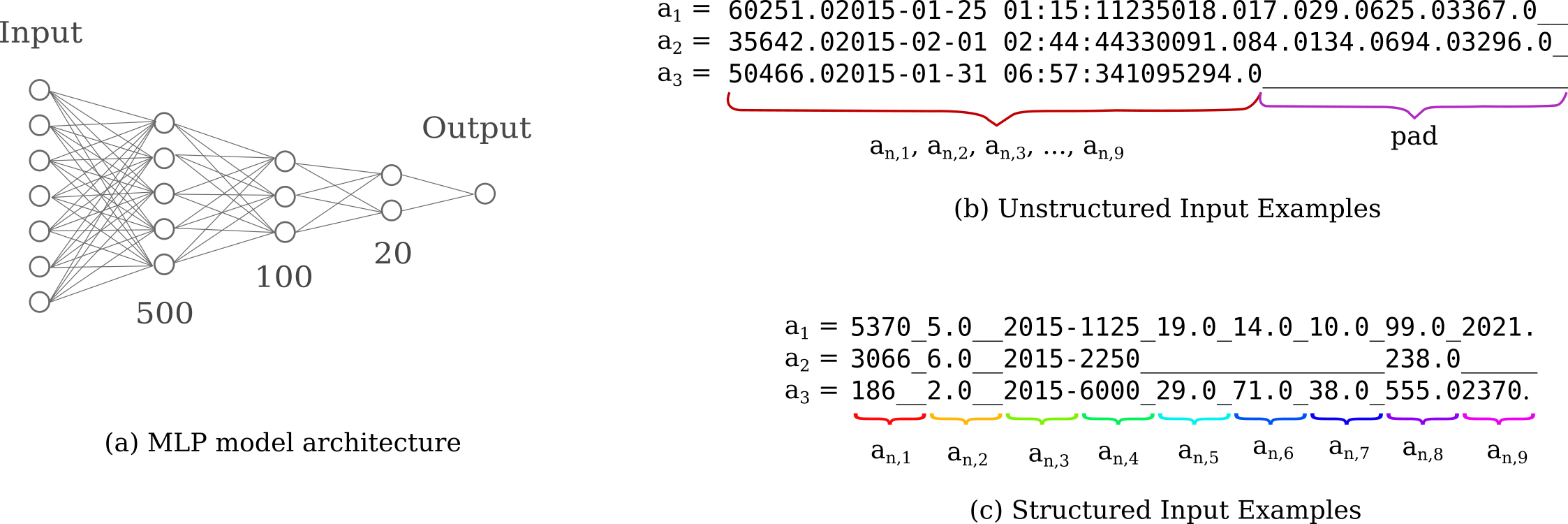}
        \caption{Encoding choices and the MLP architecture used throughout this work. In (a), layer widths for a three-hidden-layer MLP used throughout this work are denoted.}
        \label{fig1}
    \end{figure}

     Recurrent neural network models are ideally suited for modeling variable-length inputs, as they operate by examining each input element sequentially. For $x_c$ the first input to that model is $[0,1,0,0,0,0,0,0,0,0]$, and the model subsequently updates the activations for each neuron according the the weights $w$ and biases $b$ present in that network, and then perhaps generating an output which can be viewed as an observation of the activations of the final layer of neurons. If inputs are batched together for training, $x_i$ can be padded to have the same dimension of $x_c$ by appending the encoding of some given character that is not seen in the dataset to the end of the character sequence.  This method is termed `unstructured' encoding and as shown in Figure \ref{fig1} (b) for clarity.
    
    Given that the design matrix has been transformed into a sequence of characters, we first thought to implement a model commonly applied to sequences, the recurrent neural network (specifically an LSTM \citep{hochreiter1997long}).  But these models were found to be very slow to train for sequences longer than a dozen or so characters, so instead we turned to a simple fully connected neural network architecture hereafter referred to as an MLP, which is shown in Figure \ref{fig1} (a).
    
    To test the ability of deep learning models to decode character string inputs during the process of function approximation we designed three synthetically controlled outputs, $y_1, y_2, y_3$ as defined in Equation (\ref{eq4}), where $a_{n, m}$ signifies the input of row $n$ at column $m$ (for the column names for each $m$, see Figure \ref{fig6} (b)).  These are hereafter referred to as Control 1, Control 2, Control 3 respectively.
    
    \begin{equation}
        \begin{gathered}
            y_1 = 10a_{n,4} \\
             \\
            y_2 = (a_{n,3}/100)*a_{n,5} \\
             \\
            y_3 = \frac{\sin(a_{n,2})*a_{n,5}}{(a_{n,4}+0.01)} + a_{n,7} / 10 \\
            \label{eq4}
        \end{gathered}
    \end{equation}
    
    From Figure \ref{fig2}, it is clear that the MLP is capable of somewhat accurate approximation of all three control outputs with the unstructured encoding method.

\subsubsection{Input Information Preservation with Fixed-length Encodings}

    There is a significant drawback to simply skipping any missing input, and indeed there is a problem when composing variable-length input vectors from tabular datasets: as there is no prior tendency for certain characters to belong to certain columns of the input, variable-length vectors necessarily lose information on the identity of each character with respect to the feature it belongs to.  To explain why this is, imagine observing the following tensor as the first input: $[0,1,0...]$. There is no way \textit{a priori} for a model have information on which feature this input tensor corresponds to, whereas in the grid of values shown in Table \ref{table:1} one certainly would know which column (and therefore which feature) an empty value was located inside. 
    
    Therefore a notable problem with using inputs of variable length is that the information present in the design matrix as to which values belong to which input column are lost.  Unless every input element is present and all elements of a given column have the same length, a sequence-encoding method will assign different column values to different positions.
    
    We circumvent the feature identity issue in variable encodings by performing fixed-length encoding in which each column (ie feature) in the design matrix is allotted a certain number of characters.  For inputs that have fewer characters than are allotted, a placeholder character is inserted to preserve the total character number (Figure \ref{fig1}).  For example, replacing the first character from the first column in input $x_c$ with a special character gives the input $x_i$ in Equation (\ref{eq5}).
    
    \begin{equation}
        \begin{aligned}
        x_c = [[0,1,0,0,0,0,0,0,0,0,0],[0,0,1,...] ...] \\
        x_i = [[0,0,0,0,0,0,0,0,0,0,1],[0,0,1,...] ...]
        \end{aligned}
        \label{eq5}
    \end{equation}
    
    This character should not be present elsewhere in the dataset in order to avoid ambiguity in the encoding. Alternatively, one can use a zero input for missing characters, for example applied to $x_c$ makes $x_i'$ as shown in Equation( \ref{eq6}).
    
    \begin{equation}
        x_i' = [[0,0,0,0,0,0,0,0,0,0,0],[0,0,1,...] ...]
        \label{eq6}
    \end{equation}
    
    Empirically this encoding method does not lead to significantly different model performance compared to the use of a special character. 
    
    \begin{figure}[h]
        \centering
        \includegraphics[width=0.85\textwidth]{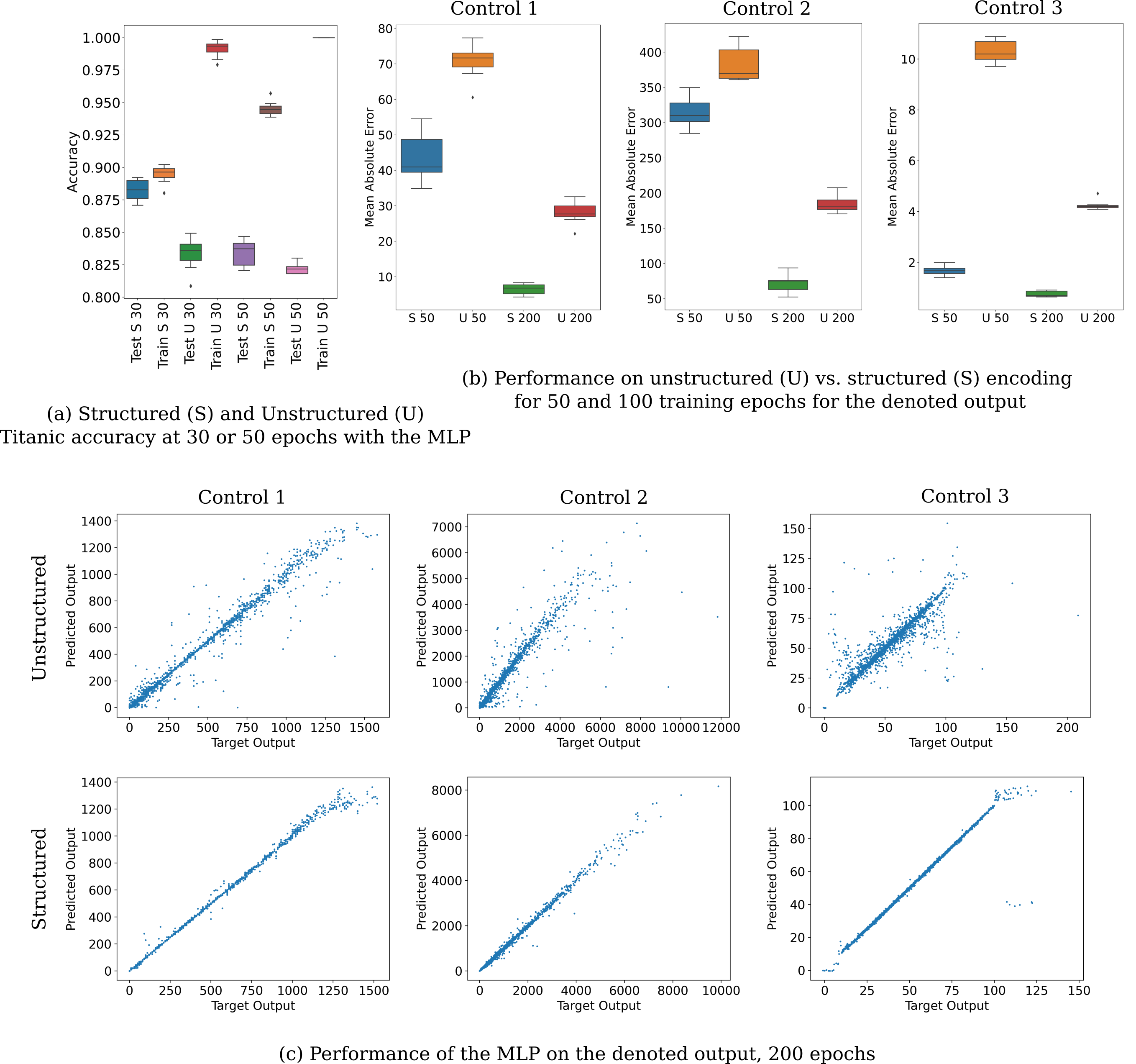}
        \caption{Accuracy of the MLP on structured and unstructured encodings of the Titanic benchmark and Control datasets, showing accuracy on test (unseen during training) data.  Structured inputs have an arbitrarily chosen number of characters per input.}
        \label{fig2}
    \end{figure}
    
    For inputs without many columns or those with few allotted characters per column, fully connected neural network architectures may be employed.  One way of representing a character sequence is to concatenate the character tensors of $x_i$ and make the resulting tensor of size $1\mathtt{x}(n*c)$, with $n$ being the number of columns and $c$ being the characters per column.  This tensor forms the first layer of the model.
    
    For longer input sequences a fully connected architecture cannot in general be used for all elements as the number of trainable parameters scales quadratically with a model of fixed size with increasing input sequence length, and for a model that scales with input size the number of parameters is exponential.  If the input size becomes prohibitive for practical use, there are in general two options: either an architecture that does not need to scale with input size may be used, or else not every character from each column may be added to make the input tensor.  We will examine the latter approach here and the former in the next section.
    
    Structured sequence encoding (with truncated inputs) may be applied to relatively small datasets successfully.  It is an interesting property of deep learning that models capable of severely overfitting data tend not to do so, and indeed when we apply the fully connected model given above to the Titanic dataset we find an average test accuracy of 83.5\% (maximum of 84.7\%) and an average training accuracy of 94.5\%, even without any regularizers present (Figure \ref{fig2}).  For reference, this accuracy is among the best of all valid entries for the Titanic challenge \citep{carlmcbrideellis2020}, which is perhaps more noteworthy when one considers that no explicit feature engineering, cleaning, preprocessing, or hyperparameter tuning has been done to achieve this result, given that we are relying on the model itself to perform these tasks implicitly.  If early stopping (at 30 epochs for all training runs rather than the randomly chosen 50 used above) is implemented to give a small amount of extrinsic regularization, we find an average of 88.2\% with the best single training run achieving 89.2\% test accuracy as shown in Figure \ref{fig2}. This exceeds what is to our knowledge the current record for all documented approaches for this dataset \citep{deotte2021}.  
    
    The same MLP is also shown to be capable of accurate function approximation on the controlled outputs, which we model as a regression task.  Regression was performed with an $L^1$ metric as the loss and used Adam \citep{kingma2014adam} optimization.  Notably, the structured sequence encoding method yielded lower test error and tended to avoid overfitting relative to unstructured input applied to the same architecture by various metrics (Figure \ref{fig2} and \ref{figs1}). Note that for all regression experiments in this work, gradient norm clipping was performed but no other explicit regularizer (dropout, layer norm etc.) was employed but despite this, accurate function approximation is consistently achieved on test (ie unseen during training) data.    
    
\section{Language model application}
    
    For longer sequences of characters, we turn to the transformer as this architecture does not necessarily need to scale in size with the input length and can therefore be used for even very long sequences. Transformers as originally described \citep{vaswani2017} were applied to problems of natural language translation and took as inputs medium-dimensional embeddings of words.  While it would be possible to embed the information of each column of a small dataset, it should also be possible to forego this step if the transformer model is capable of representing the input sufficiently well itself.  Therefore instead of embeddings we give one-hot encoded characters of our input sequence directly to the transformer encoder, with a single fully connected hidden layer used as the transformer decoder for classification or regression.  
    
    \begin{figure}[h]
        \centering
        \includegraphics[width=0.85\textwidth]{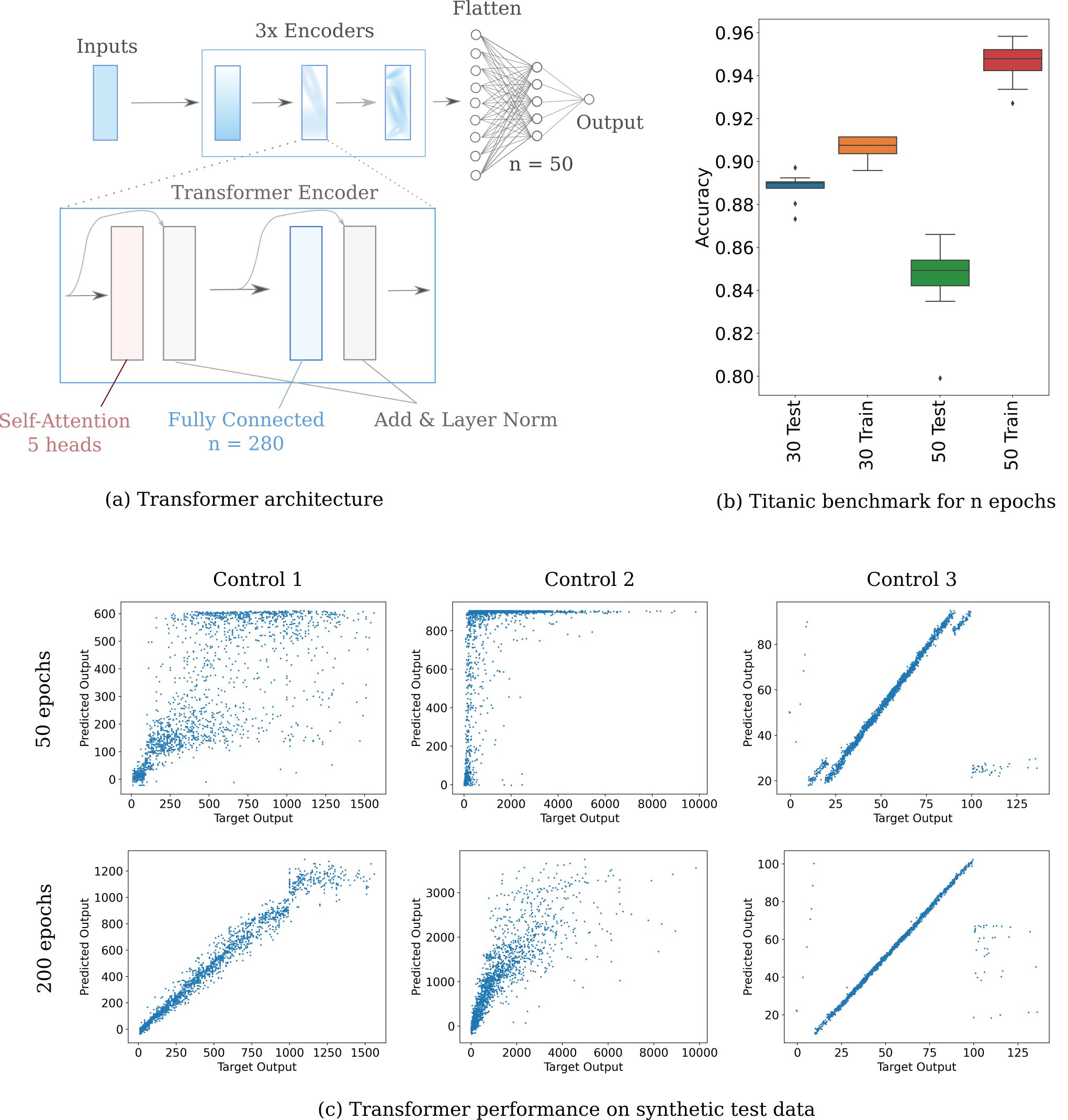}
        \caption{Transformer model performance on control and benchmark datasets. In (b) the transformer was trained for either 30 or 50 epochs, and only structured inputs were used for all experiments in this figure. In (c), all plots are of test data.}
        \label{fig3}
    \end{figure}
    
    The transformer architecture achieves slightly higher average (84.4\%) and maximum (86.6\%) test accuracy compared to the MLP on the Titanic dataset over 50 epochs (with an average training accuracy of 94.5\%) and also a higher average (88.8\%) and maximum (89.7\%) test accuracy with early stopping at 30 epochs (Figure \ref{fig3}).  On the other hand, as shown in Figure \ref{fig3} the transformer is somewhat worse than the MLP at performing function approximation regression tasks.

    Notably, positional encoding is found to not be necessary for the accuracy on regression tasks (Figure \ref{figs2}) and was not applied to tasks of classification.  Investigating why positional encoding would be superfluous, we note that contrary to what has been implied in previously published work \citep{vaswani2017} the transformer architecture does indeed supply positional information from the input.  This can be shown experimentally by permuting the input elements and observing the output: the results for this experiment are given in Figure \ref{figs2}, and it can clearly be seen that permuting the input changes the output in the general case. We note first that skip connections bypassing multi-head attention operations in the transformer encoder render the positional information in this module similar to any other fully connected architecture, and that even if these connections did not exist then self-attention key, query, and value weight vectors are unlikely to be identical such that the position of any input element may be solved for if these values are known.

\section{Abstract Sequence models form useful embeddings on their inputs}

    In the context of deep learning, an embedding is a vector space such that inputs of greater `similarity' in the input space also have a higher similarity in that vector space, which is usually of lower dimensionality than the input.  The `similarity' metric on the input is usually application-specific: for example, when embeddings of English words are formed the meaning of `similarity' is almost identical to the usual English term (ie dog and hound are more `similar' than dog and cat).  The output similarity is usually a metric on the embedding vector space.  
    
    Throughout this work we have fed characters to deep learning models with no prior embeddings, meaning that the character string encodes no other information than the identity of each character.  The successes of models applied to classification and regression tasks suggests that they are capable of forming their own embeddings of the input, as it pertains to addressing the task defined by the loss function.
    
\begin{figure}[h]
        \centering
        \includegraphics[width=0.98\textwidth]{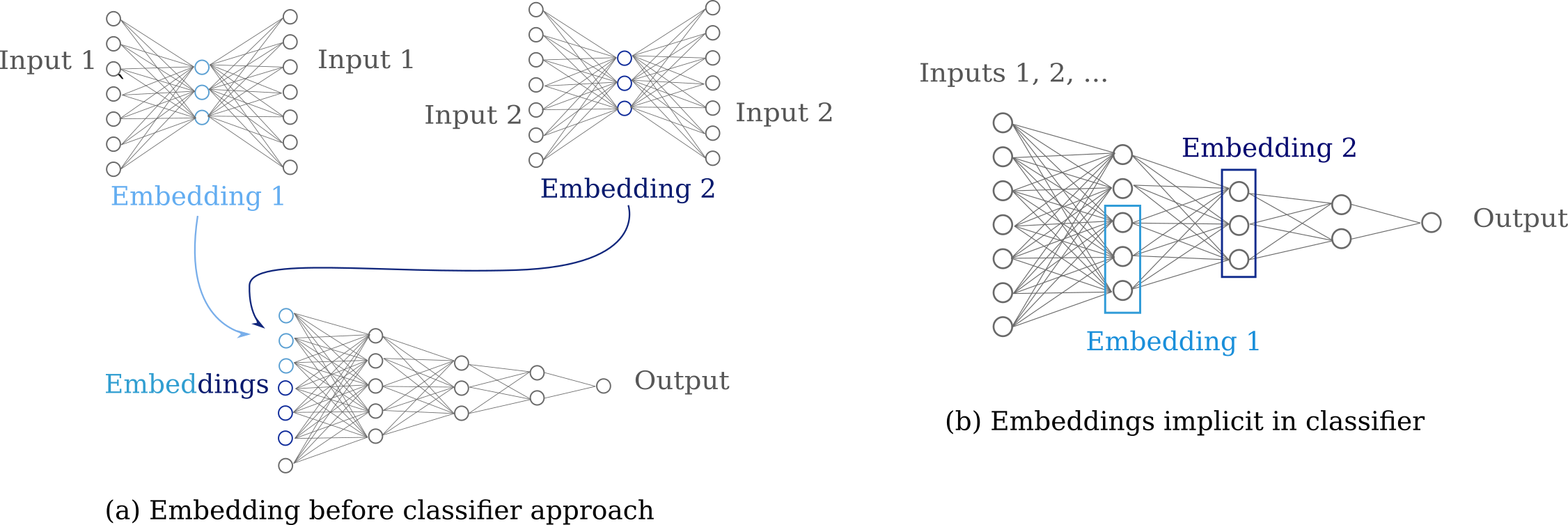}
        \caption{Theory of embeddings}
        \label{fig4}
    \end{figure}

    This is a fundamentally different approach than first forming embeddings of the input and subequently compiling an input to a classifier or regression model from these embeddings. For natural language modelling, the embedding is normally obtained using an autoencoder's latent space to reduce the dimensionality of the input (which can be tens or hundreds of thousands of possible words in a dictionary) to a vector in a space with far fewer dimensions, usually in the ballpark of 512.  But given that there can only be at most hundreds of possible characters in any given sequence, we reasoned that this pre-processing step should be unnecessary if one assumes that the model forms a meaningful representation of an input with respect to the task at hand.  The difference between these approaches is illustrated for clarity in Figure \ref{fig4}.

    The ability of a model to form an embedding can be observed directly in the case where the model is trained for regression, as one can assign a metric of choice to compare distances between hidden layers activations.  The assumption behind observing an embedding distance is that hidden layers with `more similar' activations (as measured by some metric) are likely to have `more similar' inputs with respect to the learned task. Here we define an $L^1$ metric on the layer outputs (activations) given input $a_n$ and model parameters $\theta$ by concatenating all elements of that output into a single index $i$ before summing the absolute values of the corresponding element differences (\ref{eq7}). This is analogous to an $L^1$ version of the Frobenius norm.
    
    \begin{equation}
        m_{1, 2} = \sum_i |O(a_1, \theta)_i - O(a_2, \theta)_i|
        \label{eq7}
    \end{equation}
    
\begin{figure}[h]
        \centering
        \includegraphics[width=0.98\textwidth]{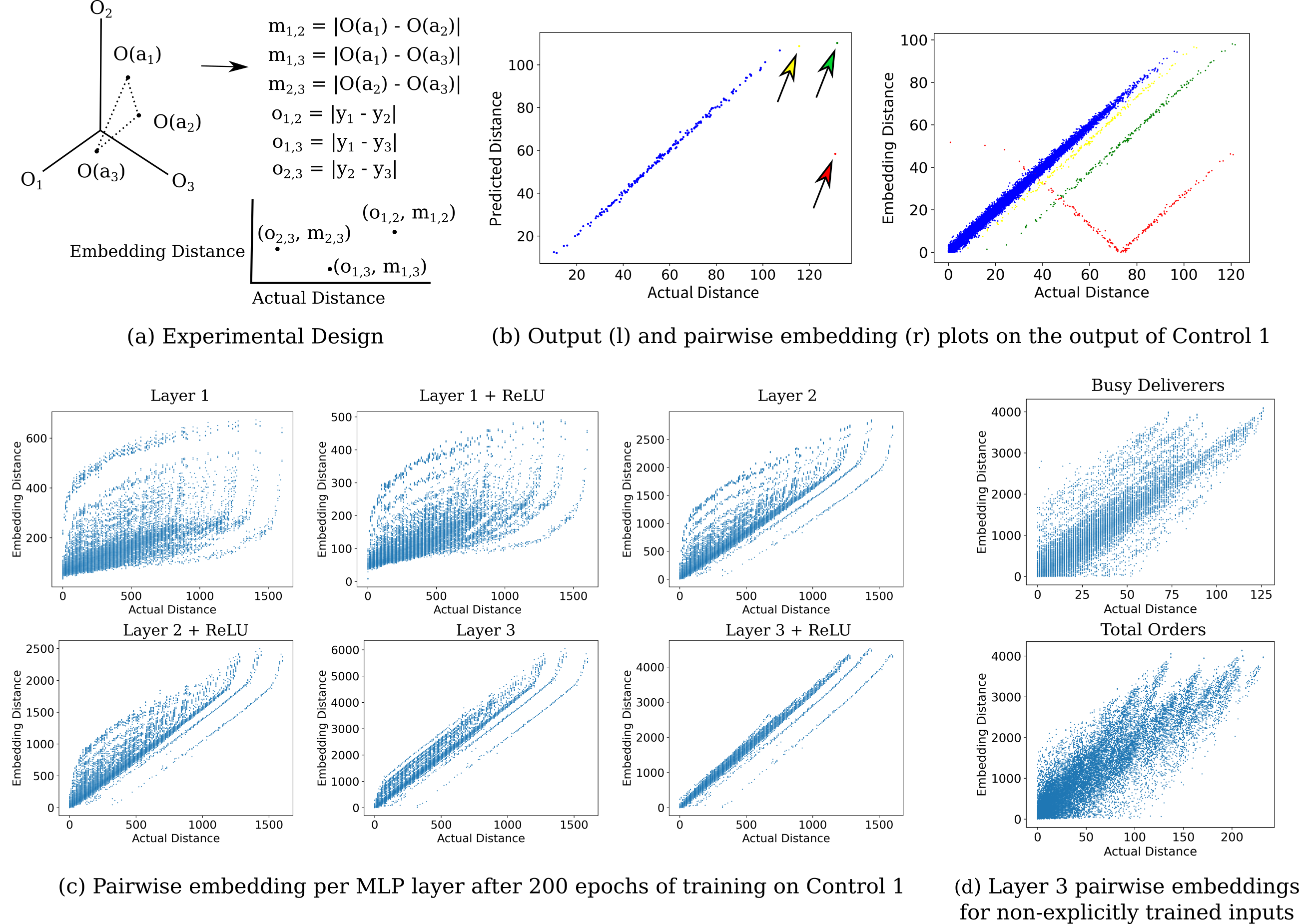}
        \caption{Sequence models form useful embeddings of the input in their hidden layers. (a) Pairwise embedding distance approach: $L^1$ metric of the difference between all pairs of points in the output space $O$ is obtained and regressed on the actual difference (defined on the input). (b) An example comparison between the predicted versus actual accuracy regression plots used elsewhere in this work to the pairwise embedding distance versus output distance (or equivalently one input element distance). (d) A model trained to predict Control 1 (which is defined using only $a_{n,4}$) also embeds $a_{n,5}$ and $a_{n,6}$. All embeddings shown are made from test data using the pairwise approach.}
        \label{fig5}
    \end{figure}

    We sought to replicate the intuitive ease that the plots of the predicted output $O(a_n, \theta)$ versus ground truth $y$ values that previous figures offered. It is generally difficult to directly visualize vector spaces of more than three or four dimensions, however, so we performed dimensionality reduction on the embedding space by observing the measures $m_{1, 2}, m_{1, 3}, m_{1, 4}, ..., m_{n, o}$ of all unique pairs of a subset of points in that space according to (\ref{eq7}) and regressing on the corresponding distance between outputs as shown in Figure \ref{fig5} (a).  As this pairwise comparison technique is not to our knowledge a standard way of visualizing high-dimensional spaces, we first compare the plot of prediction accuracy of 200 test elements of an MLP trained on Control 1 to a plot of all distances of pairs of points for those same elements to gain familiarity with this technique when applied to outputs of one element.  From the example in Figure \ref{fig5} (b) it is clear that each point that lies a noticeable distance from the line $y=x$ (ie a perfect prediction) forms a cluster of points in the pairwise comparison plot. This cluster is near-linear or piecewise linear being that nearly all other points lie close to $y=x$.  
    
    The pairwise technique similarly reduces points in many-dimensional output space of hidden layers to points in two dimensions to form an analogue of prediction accuracy for those high-dimensional spaces.  This transformation is non-unique such that many possible arrangements of points in embedding space may lead to identical pairwise comparison plots.  But as embeddings are in general defined on relative position only (ie which elements are closer and which are further) non-invertibility is not a problem for visualization.  A useful embedding is one in which there is a correlation between the embedding space distance and the output distance (or equivalently in this case the input distance).

    As can be observed in Figure \ref{fig5} (c) and (d) for Control 1 and Figure \ref{figs3} for Control 3, most hidden layers of the MLP model form embeddings of the input with respect to addressing the task defined by the loss function and desired output, which in this case is a function on one input field. But it is important to note that the model also forms embeddings on input fields that are not strictly necessary for addressing the defined task (Figure \ref{fig5} (d).  Given that most applicable tasks defined on tabular data are not likely to require knowledge of only one of the input fields, one can with reasonable certainty assume that at least a few of the most important input features (important for the defined task, that is) are embedded in the hidden layers of a trained model.
    
    The pairwise embedding measurement technique also gives insight into why unstructured sequences do not perform as well as structured ones: an MLP trained on Control 3 shows that although there exists a correlation between embedding distance (of the third layer) and output distance, there is no distinction between separate inputs in that embedding (Figure \ref{figs3}).  This can be compared to the structured sequence encoding where each outlier's cluster of points is easily identifiable, suggesting that indeed the MLP performs better with structured sequences because such models can learn to identify each unique input.  
    
    Likewise, when we investigated embedding present in the transformer encoder trained on Control 1, the encoder's last fully connected layer contain rather poor embeddings of the input (Figure \ref{figs3}). It is interesting to note that the embedding results suggest that most inputs are indeed mapped to approximate a manifold of one dimension with respect to the desired output.  To see this, observe that a two- or more-dimensional manifold would not in general yield a linear approximation in the pairwise distance plot because there are expected to be many points that exist on a hypersurface that are equidistant from one given point (ie for any point above a plane $p$ there is a circular region where all planar points are equidistant from $p$) that would not be equivalent in the output space.  
    
    There are two noteworthy addendums to this embedding measurement method: firstly, correlation between pairwise embedding distance and output (or input) scalar distance implies a learned embedding has formed but the lack of a correlation does not necessarily mean that an embedding has not been obtained.  This can be seen in Figure \ref{figs3} (d), where there is little sign of an embedding formed in a trained transformer encoder's final linear layer on Control 3 data. Despite this seeming lack of embedding in the transformer encoder, when one compares the output accuracy of models with and without this module it is clear that the encoder indeed organizes the input information in some way such that the final layer is better able to perform a regression.
    
    Secondly, it should be recognized that using the embeddings implicit in a classification or regression model is not always appropriate: an embedding learned via an autoencoder or other unsupervised technique likely obtains information on this as well as other aspects of the input data variation that are not relevant to predicting the output $y$.  For the case where many inputs may be important for an output, it is more likely that a hidden layer of a supervised model contains more broad information being that the model cannot be sure of what is and is not important for the task.

\section{Input attribution for sequence models}

    For many machine learning tasks, it is useful to be able not just to arrive at some prediction but also to understand how that prediction was arrived at.  For linear models this process is fairly straightforward, as for example given a multiple linear regression we have only to observe the weights associated with each input element to determine which is or is not important for a prediction.  Simply observing the weights directly can give this information because linear models are additive and scaling, meaning that a change in one variable does not change the model's other variables.
    
    This is not the case for nonlinear models, which in general have no such easy interpretations available from their parameter space.  This is because the effect of one parameter may depend on the precise values of the other parameters as nonlinear models are in general non-additive.  Instead of attempting to understand how a deep learning model arrives at an output via its parameters, therefore, we instead focus on the relationship between the model's inputs and its output. This can be done by simply covering up (occluding) certain elements of the input and comparing the output to the original, or else by using the differentiability of deep learning models to find the gradient of the output with respect to the input as if the input were a trainable parameter.
    
\subsection{Attribution with Occlusion}
    
    Occluding an input may be accomplished by simply removing individual elements before observing the effect on the output.  One way to accomplish this is to generate an occluded input $x_o$ by simply replacing each element in turn with a special character, $u$ as shown in Equation (\ref{eq8}). The full occlusion process repeats this replacement for each character in the input. It is important to occlude with a different character than any that are seen in the original dataset in order to prevent an occlusion from introducing unwanted information, which could lead to erroneous attribution scores.
    
    \begin{equation}
        x_o = f(u \;2015-02-06 \; 22:24:17 \;1845 ...)  
        \label{eq8}
    \end{equation}
    
    For regression outputs, the occlusion value $v$ by taking the absolute value of the difference between the outputs of the model given the occluded input $x_o$ and the output given the complete input $x_c$ (\ref{eq9}) and this can be extended to classification tasks with many outputs by performing an $L^1$ metric measurement (\ref{eq10}) where $i \in C$ denotes a category in the set of all possible output categories $C$.
    
    \begin{equation}
        v = \vert O(x_c, \theta) - O(x_o, \theta) \vert
        \label{eq9}
    \end{equation}
    
    \begin{equation}
        v = \sum_i \vert m(x_c)_i - m(x_o)_i \vert
        \label{eq10}
    \end{equation}

    With image data it is often necessary to occlude more than one pixel at a time, and for sequence data we experimented with occluding between 1 and 4 characters at once.  For a 2-character occlusion (with a stride of 1) followed by a maximum normalization in which each occlusion value is divided by the maximum of all values, models trained on Control 1 yields consistent attribution measurements for various inputs (Figure \ref{fig6}).

\subsection{Attribution with Gradient*Input}

    Neural networks are trained via gradient descent and are thus necessarily differentiable.  Because of this, one can find the gradient of the output with respect to the input and use this information to understand how a small change to each input element would change the output. 

    When applied to image data in which each pixel (or in other words each input element) can be expected to have a non-zero value, we would not expect to lose much information with Hadamard multiplication and furthermore the process is fairly intuitive: brighter pixels (ie those with larger input values) that have the same gradient values as dimmer pixels are considered to be more important. 

    With one-hot character encodings a similar approach may be made: the absolute value of the gradient of the output with respect to the input followed by Hadamard multiplication of that input (\ref{eq11}). 
    
    \begin{equation}
        v = \lvert \nabla_x O(x, \theta) \rvert * x
        \label{eq11}
    \end{equation}
    
    As for occlusion, this method followed by max norm yields consistent attribution over many inputs (Figure \ref{fig6}).
    
    \subsection{Combined Importance}
    
    Gradient*Input and Occlusion both address the question of what would happen to a model's prediction if some input value changed.  Furthermore, these are somewhat complimentary methods as occlusion measures the output given a discrete change whereas gradient*input predicts the change in output given an infinitesimal change.  As deep learning models are often approximations of functions that are by no means simple or convex, it is usually the case that the measure of importance assigned by occlusion is different from the one assigned by gradient*input due to poor correspondence between local and global structure in that function.
    
    This is motivation to combine these two measurements into one attribution score.  First the average of the max-normalized occlusion $v_o$ and gradient*input $v_g$ is found for each input element, and then all elements per given input column $i$ (with indices $j$ of elements in that column) are combined via the maximum value per column as shown in Equation (\ref{eq12}).
    
    \begin{equation}
        v_i' = \mathtt{max} \; ((v_o)_{i,j} + (v_g)_{i,j})/2
        \label{eq12}
    \end{equation}
    
\begin{figure}[h]
        \centering
        \includegraphics[width=0.95\textwidth]{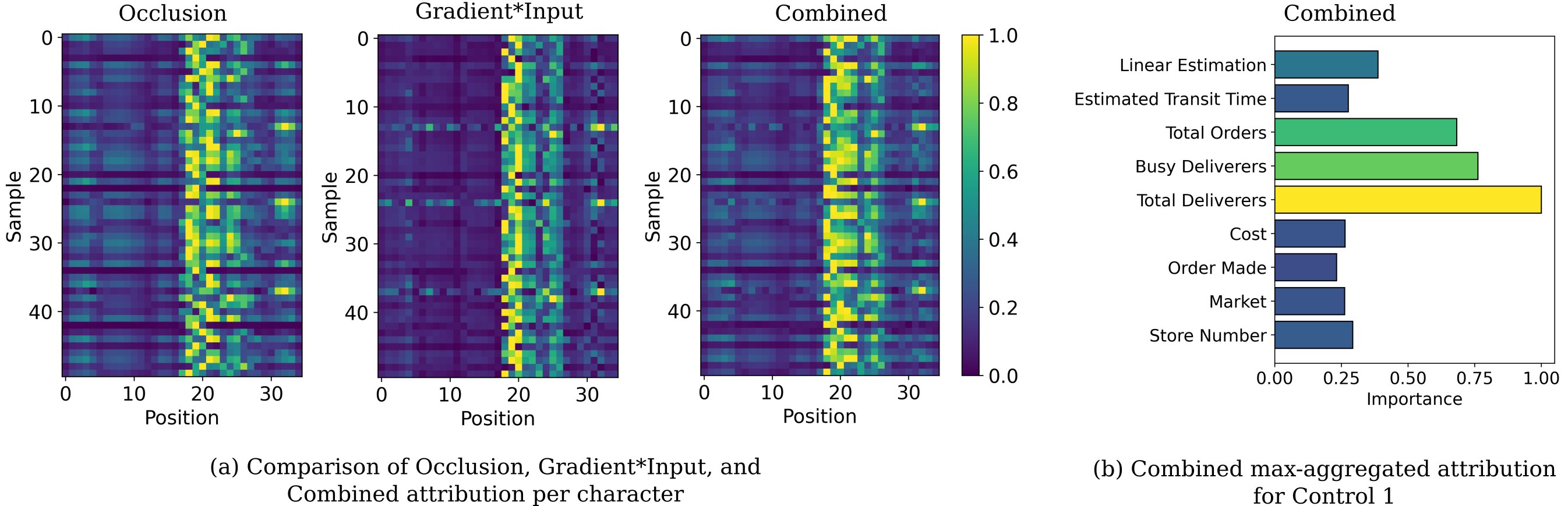}
        \caption{Input attribution for interpretation.  In (b), inputs are in order from bottom to top, ie `Store Number' is equivalent to $a_{n,0}$ and `Linear Estimation' is column $a_{n,8}$. Note that the correct argument max for input attribution is found.}
        \label{fig6}
    \end{figure}

\begin{figure}[h]
        \centering
        \includegraphics[width=0.85\textwidth]{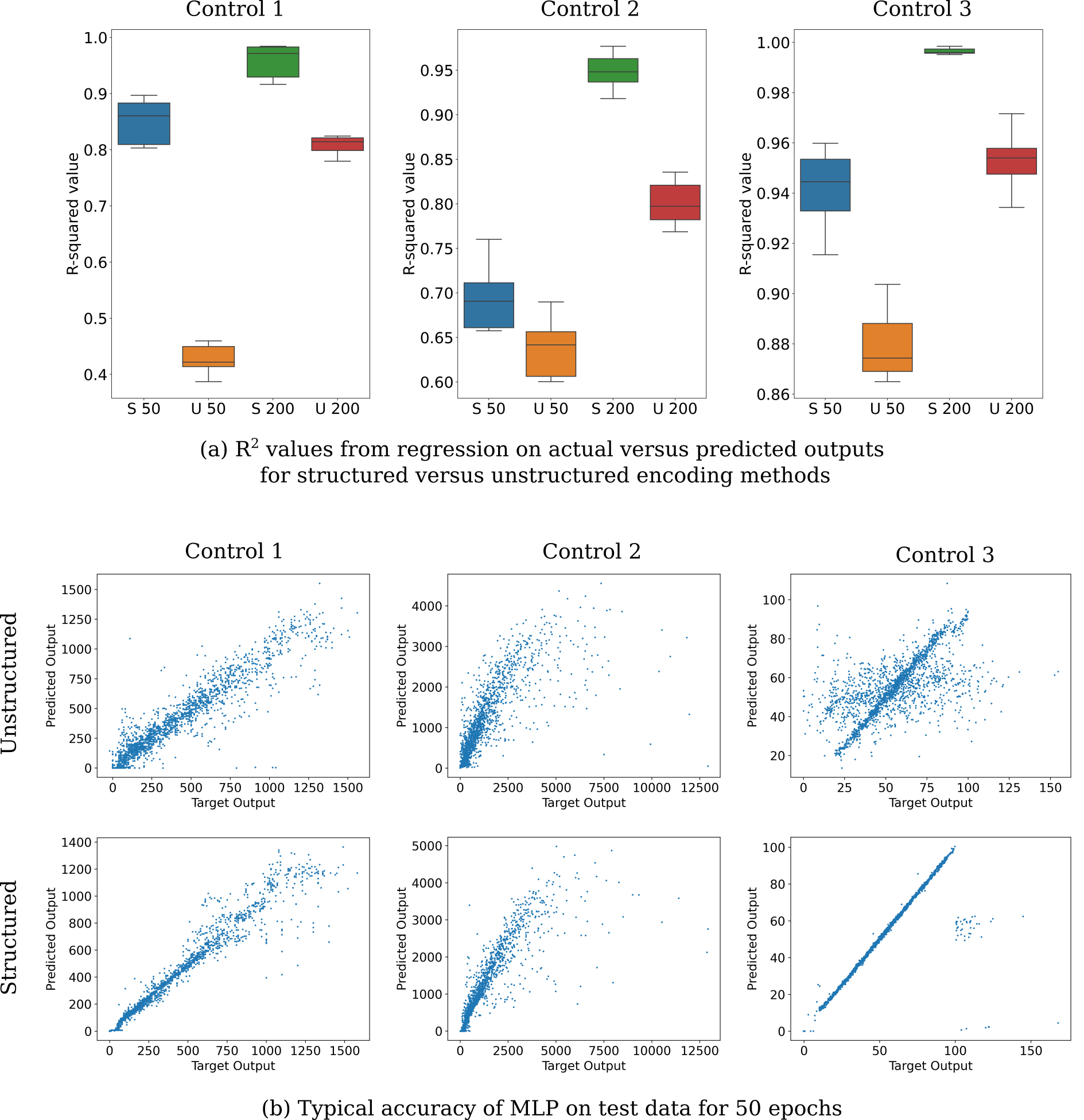}
        \caption{MLP performance on deterministic and nondeterministic datasets continued}
        \label{figs1}
    \end{figure}

\begin{figure}[h]
        \centering
        \includegraphics[width=0.95\textwidth]{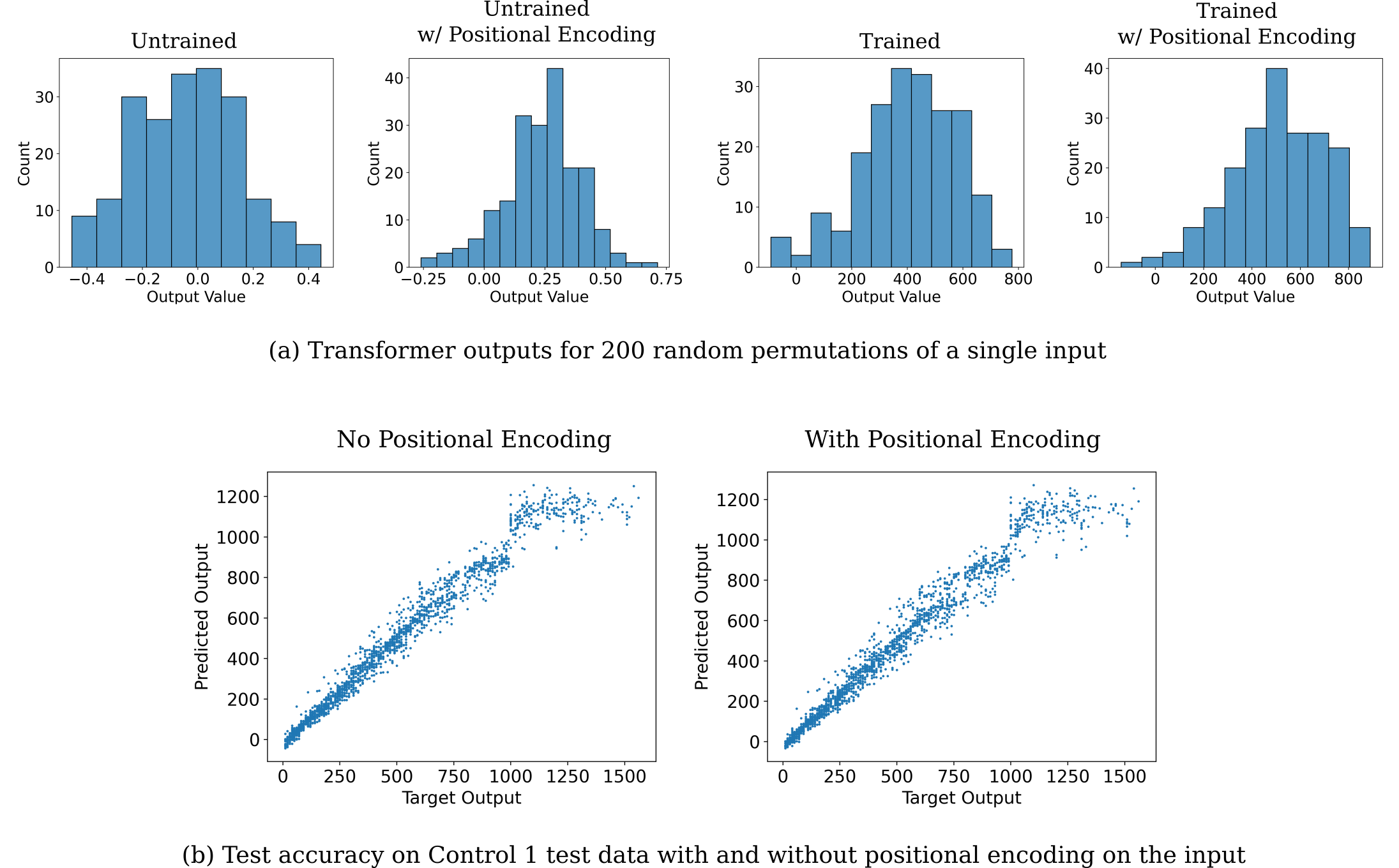}
        \caption{Transformer encoders transmit positional information. Transformer accuracy for function approximation with and without positional encoding.}
        \label{figs2}
    \end{figure}

\begin{figure}[h]
        \centering
        \includegraphics[width=0.85\textwidth]{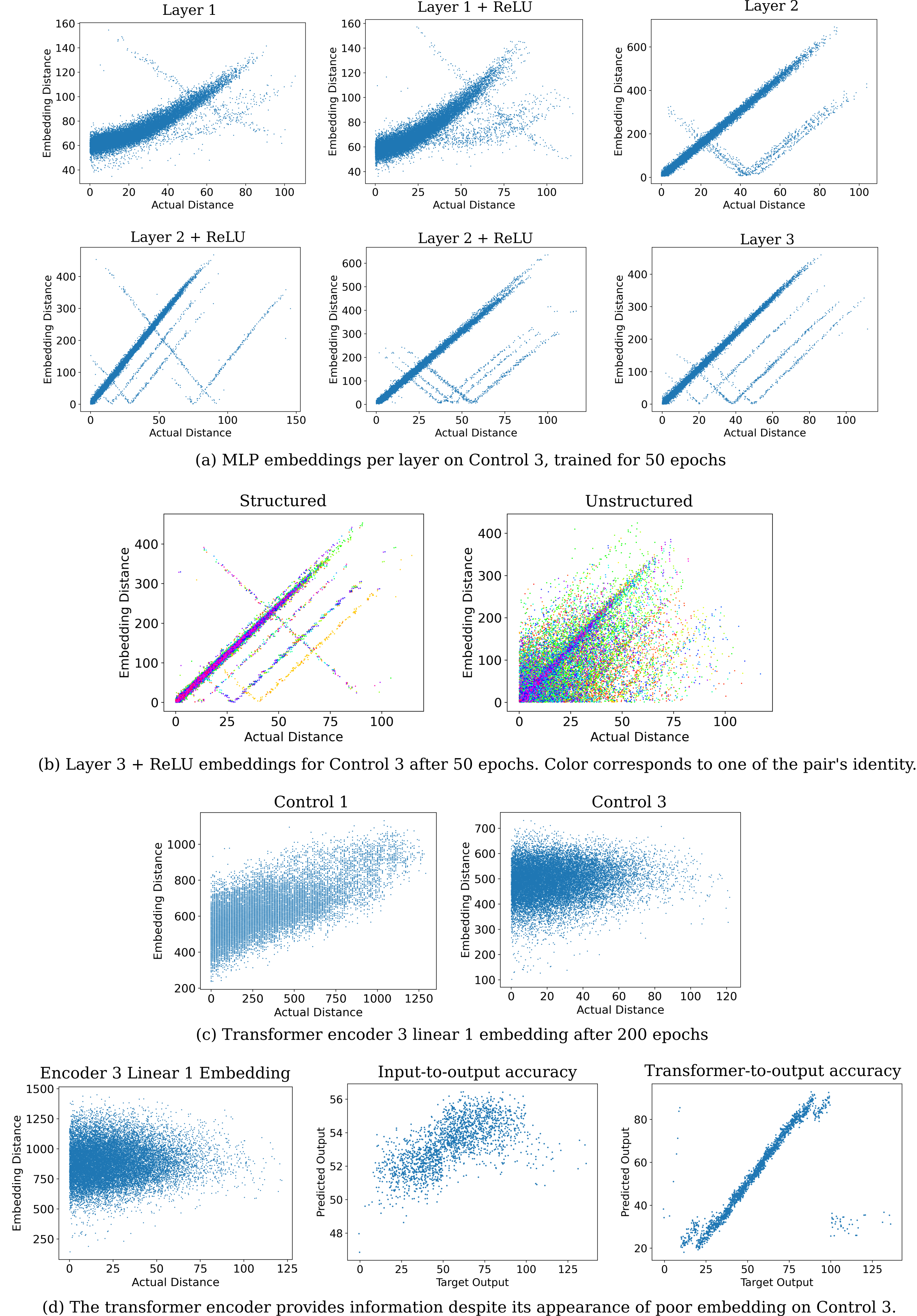}
        \caption{Embeddings continued. In (d) the hidden layer following the transformer encoder is removed such that the outputs follow a linear transformation from the encoder. The encoder's embedding appears to be poor via the pairwise distance method but training a model without the encoder severely decreases regression accuracy.}
        \label{figs3}
    \end{figure}

    The results for combined importance are shown in Figure \ref{fig6}.  After training most inputs have highest importance correctly placed on the characters that determine the output of the linear control dataset.   
    
\begin{figure}[h]
        \centering
        \includegraphics[width=0.85\textwidth]{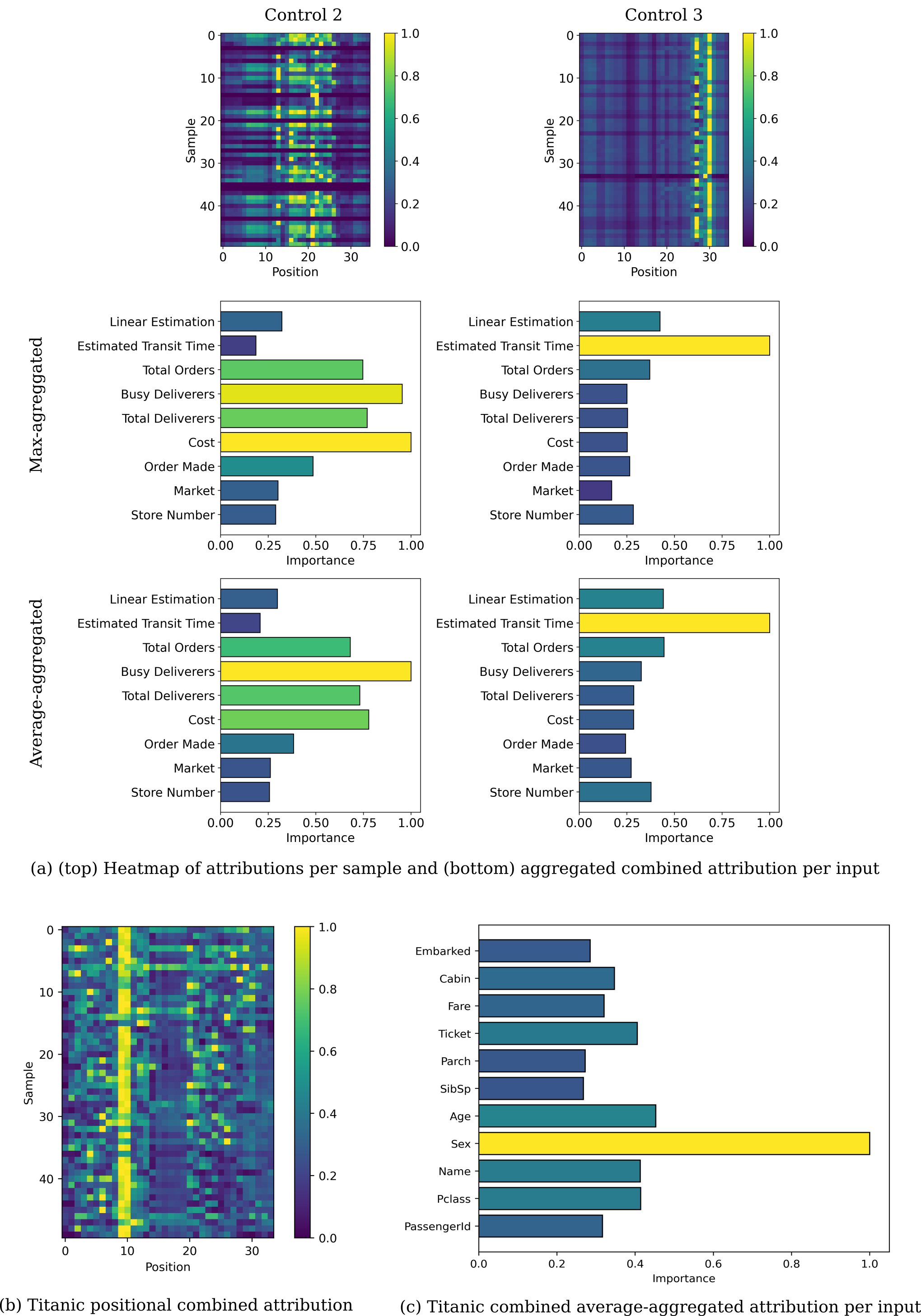}
        \caption{Input attributions continued.  (a) Max- and average- aggregated attribution per input field are similar for Controls 2 and 3.  Inputs are in order from bottom to top, ie `Store Number' signifies $a_{n,0}$ and `Linear Estimation' signifies $a_{n,8}$.  (b) and (c), Titanic dataset input attributions.}
        \label{figs4}
    \end{figure}

    It is often useful to obtain information on per-column importance from many inputs at once rather than from only one at a time.  If this is the case, we can average together the combined attributions obtained in (\ref{eq10}) over many inputs. Attribution values are aggregated per column and averaged over 100 samples are shown in Figures \ref{fig6} and \ref{figs4}.
    
    One notable benefit of the ability to ascribe average input attribution per field is that it allows for a rough estimation of which inputs can be expected to have meaningful embeddings in a model's hidden layers.  Given some task and a model, a higher attribution for some input field $m$ such that $a_{n,m}$ corresponds to a higher likelihood of that model transmitting the information relevant for the task from $a_n$ to the output.  Assuming no skip connections, each hidden layer must be able to represent $a_{n,m}$ in such a way that this information is not lost.  
     
\section{Implications}

    This work provides a proof-of-concept for the idea that steps of the cleaning, formatting, and otherwise preprocessing that normally must be undertaken for a dataset destined for a machine learning model may be obviated by a simple encoding scheme that allows for directly feeding data to a deep representational model.  Evidence is also presented to show that the hidden layers of a regression model are capable of forming embedding of the input, showing that embedding as a pre-processing step may also be omitted.
    
    Besides the utility of relying on input representation to bypass data preprocessing, it has been shown that the representations formed during the learning process are generally superior to those formed by explicit feature engineering in the case of the Titanic dataset.  This finding is not surprising given that it has been observed in the context of chess engines \citep{silver2017} and image recognition \citep{olah2017feature} among others: the representations formed by deep learning models are often unintuitive for a human to follow, and are rarely arrived at with manually chosen rules.  But the ability of these learned representations to outperform rule- or judgement-based practices provides a basis for viewing these representations as important even if they would not be chosen via classical inference.

\bibliographystyle{unsrtnat}
\bibliography{references} 

\beginsupplement

\section{Appendix}

    All figures (unless specifically noted otherwise) show results on test data. It should be noted that the dataset for Controls 1 - 3 was shuffled before splitting into training and test sets, making the test set's specific values change between training runs.  Plots of Titanic accuracy are Tukey-style box plots where the box showns the IQR and the center line denotes the distribution's median.

\end{document}